%% file: main.tex
\documentclass[10pt,twocolumn,letterpaper]{article}
\usepackage{bm}
\usepackage[hyphens]{url}
\usepackage[shortlabels,inline]{enumitem}
\usepackage{multirow}
\usepackage{bbding} 
\usepackage{pifont} 
\usepackage{amsthm,amsmath,amssymb}
\usepackage{mathrsfs}
\usepackage{float}
\usepackage[utf8]{inputenc} 
\usepackage[T1]{fontenc}    
\usepackage{booktabs}       
\usepackage{amsfonts}       
\usepackage{nicefrac}       
\usepackage{microtype}      
\usepackage{xcolor}         
\usepackage{multirow}
\usepackage{graphicx}
\usepackage{xspace}
\usepackage{balance}
\usepackage{colortbl}
\usepackage{scrextend}

\usepackage[accsupp]{axessibility}

\definecolor{cvprblue}{rgb}{0.21,0.49,0.74}
\definecolor{fakerefcolor}{rgb}{0.21,0.49,0.74}

\usepackage[pagebackref,breaklinks,colorlinks,allcolors=cvprblue]{hyperref}

\usepackage[pagenumbers]{cvpr}        

\input{sections/math_cmds.tex}

\def\eg{\textit{e.g.}}
\def\ie{\textit{i.e.}}

\def\etc{\textit{etc.}}

\newcommand{\best}{\cellcolor{blue!25}}

\newcommand*{\mcs}{\includegraphics[scale=0.045]{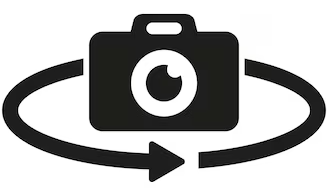}}%
\newcommand*{\cs}{\includegraphics[scale=0.05]{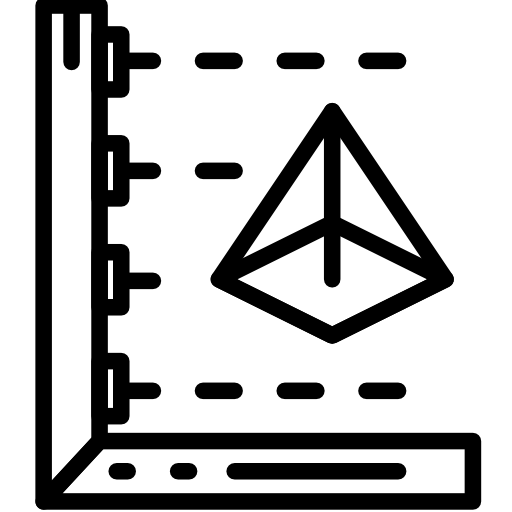}}%
\newcommand*{\ds}{\includegraphics[scale=0.025]{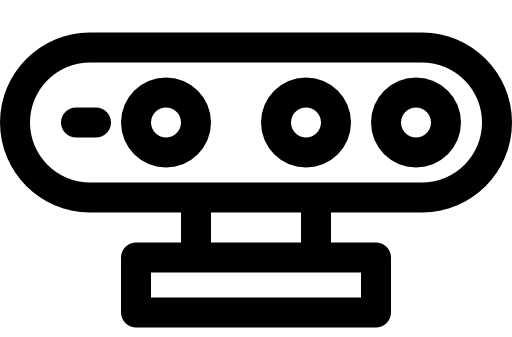}}%
\newcommand*{\imu}{\includegraphics[scale=0.045]{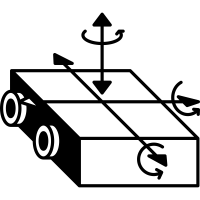}}%
\newcommand*{\pa}{\includegraphics[scale=0.04]{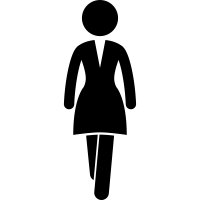}}%

\usepackage[capitalize]{cleveref}
\crefname{section}{Sec.}{Secs.}
\crefname{table}{Tab.}{Tabs.}
\crefname{figure}{Fig.}{Figs.}
\newcommand{\datasetname}{WildAvatar\xspace}


\def\paperID{00235} 
\def\confName{CVPR}
\def\confYear{2025}

\title{\textsl{\datasetname}: Learning In-the-wild 3D Avatars from the Web}

\author{%
    Zihao Huang$^{1, 2}$, Shoukang Hu$^2$, Guangcong Wang$^{3}$, Tianqi Liu$^{1, 2}$,  \\
    Yuhang Zang$^4$, Zhiguo Cao$^1$, Wei Li$^{2, \dagger}$, Ziwei Liu$^2$
\vspace{0.3em} \\
  $^1$School of AIA, Huazhong University of Science and Technology, \\
  $^2$S-Lab, Nanyang Technological University, $^3$Great Bay University, $^4$Shanghai AI Laboratory 
    \vspace{0.2em} \\
  \href{https://wildavatar.github.io/}{https://wildavatar.github.io/}
    \vspace{-5pt}
}

\makeatletter
\def\thanks#1{\protected@xdef\@thanks{\@thanks
        \protect\footnotetext{#1}}}
\makeatother

\begin{document}

\twocolumn[{%
\renewcommand\twocolumn[1][]{#1}%
\maketitle
\vspace{-30pt}
\begin{center}
    \centering
    \includegraphics[width=0.99\linewidth]{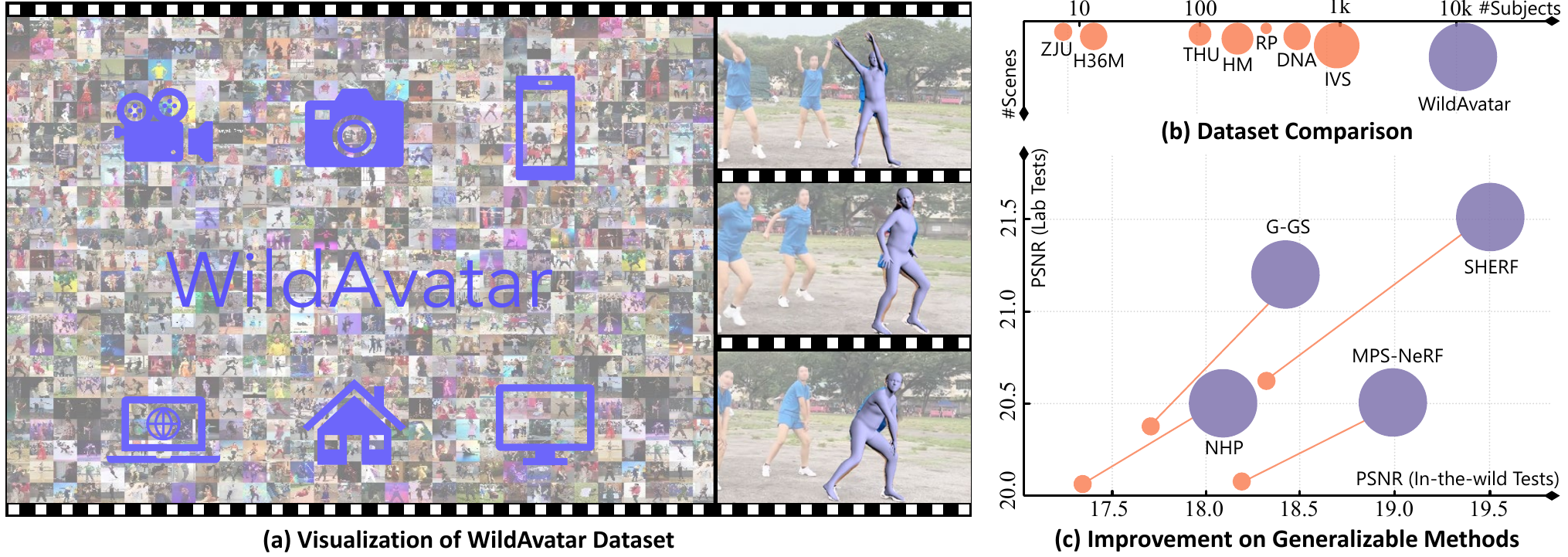}
\vspace{-5pt}
    \captionof{figure}{Overview of \datasetname. (a) Unlike previous laboratory datasets for 3D avatar creation, \datasetname curates in-the-wild web videos. (b) With $10k+$ human subjects and scenes, \datasetname is at least $10\times$ richer than the previous datasets. (c) It contains high-quality annotations and demonstrates impressive potential to boost the quality and generalizability of avatar-creation methods.}
   \label{fig:teaser}
\vspace{-5pt}
\end{center}%
}]

\input{sections/0_abstract}

\thanks{$\dagger$: Corresponding author.}

\input{sections/1_introduction}

\input{sections/2_related_work}

\input{sections/3_method}
\input{sections/4_dataset}

\input{sections/5_experiments}

\input{sections/6_conclusion}

{
    \small
    \bibliographystyle{ieeenat_fullname}
    \balance
    \bibliography{main}
}

\clearpage
\appendix
\input{sections/7_appendix}

\end{document}

%% file: sections/math_cmds.tex
\usepackage{amsmath,amsfonts,bm}

\def\eqref#1{equation~\ref{#1}}

\def\1{\bm{1}}

\DeclareMathAlphabet{\mathsfit}{\encodingdefault}{\sfdefault}{m}{sl}
\SetMathAlphabet{\mathsfit}{bold}{\encodingdefault}{\sfdefault}{bx}{n}



%% file: sections/0_abstract.tex
\begin{abstract}
Existing research on avatar creation is typically limited to laboratory datasets, which require high costs against scalability and exhibit insufficient representation of the real world.
On the other hand, the web abounds with off-the-shelf real-world human videos, but these videos vary in quality and require accurate annotations for avatar creation.
To this end, we propose an automatic annotating pipeline with filtering protocols to curate these humans from the web.
Our pipeline surpasses state-of-the-art methods on the EMDB benchmark, and the filtering protocols boost verification metrics on web videos.
We then curate \textbf{\textsl{\datasetname}}, a web-scale in-the-wild human avatar creation dataset extracted from YouTube, with $10,000+$ different human subjects and scenes. \datasetname is at least $10\times$ richer than previous datasets for 3D human avatar creation and closer to the real world.
To explore its potential, we demonstrate the quality and generalizability of avatar creation methods on \datasetname. 
We will publicly release our code, data source links and annotations to push forward 3D human avatar creation and other related fields for real-world applications.
\end{abstract}

%% file: sections/1_introduction.tex
\section{Introduction}
\label{sec:intro}

 \begin{table*}[!t]
  \centering
  \setlength{\tabcolsep}{8pt}

  \begin{tabular}{cccc|cccc}
    \toprule
    \textbf{Dataset} & \#\textbf{Sub.}/\textbf{Sce.} & \textbf{Type} &  \textbf{Cost} &
    \textbf{Dataset} & \#\textbf{Sub.}/\textbf{Sce.} & \textbf{Type} &  \textbf{Cost} \\
    \midrule
    
ZJU-Mocap~\cite{neuralbody}        
& 9/6  & Lab & \mcs+\pa & 
RenderPeople~\cite{renderpeople}    
& 482/$-$ & Lab & \cs+\pa \\
HuMMan~\cite{humman}                
& 339/20 & Lab  & \mcs+\ds+\pa &
THuman~\cite{thuman}                
& 100/10 & Lab  & \ds+\pa  \\
DyMVHumans~\cite{PKU-DyMVHumans}                   
& 32/45 & Lab  & \mcs+\pa &
THuman2.0~\cite{thuman2}            
& 500/$-$ & Lab  & \mcs+\ds+\pa  \\
Human3.6M~\cite{h36m}           
& 11/15  &  Lab &  \mcs+\ds+\pa &
THuman3.0~\cite{thuman3}            
& 20/$-$ &  Lab & \mcs+\ds+\pa    \\
THuman4.0~\cite{thuman4}            
& 3/3   &   Lab & \mcs+\pa    &
THuman5.0~\cite{thuman5}             
& 10/10 &  Lab &  \mcs+\pa \\
Hi4D~\cite{Hi4D}              
& 40/10  & Lab &  \mcs+ \cs+\pa   &
DNA~\cite{DNA-Rendering}
& 500/1187 &  Lab & \mcs+\ds+\pa    \\
DynaCap~\cite{DynaCap}
&  4/5  &  Lab & \mcs+\pa  &
MultiHuman~\cite{multihuman}
&   50/$-$  &  Lab & \mcs+\pa   \\
UltraStage~\cite{UltraStage}
&  100/20  &  Lab &  \mcs+\pa   &
Actors-HQ~\cite{HumanRF}
&  8/21  &  Lab & \mcs+\pa  \\
HUMBI~\cite{HUMBI}
&   772/4  &  Lab & \mcs+\pa       &
NHR~\cite{NHR}
&   4/3   &  Lab & \mcs+\pa     \\
AIST++~\cite{AIST}
&   30/10   &  Lab & \mcs+\pa       &
MPII~\cite{MPII}          
&    8/1   &  Lab &  \mcs+\pa     \\
ENeRF~\cite{ENeRF}
&   7/4   &  ITW &  \mcs+\pa        &
3DPW~\cite{3dpw}          
&    7/4   &  ITW &  \mcs+\imu+\pa    \\
FreeMan~\cite{freeman}
&    40/123   &  ITW &   \mcs+\pa       &
SynWild~\cite{Vid2Avatar}        
& 5/5 & ITW & $--$  \\
NeuMan~\cite{neuman}        
& 6/6 & ITW &  $--$  &
TikTok~\cite{TikTok}        
& 340/340 & ITW & $--$  \\
IVS-Net~\cite{ivsnet}       
& <700/<700  & ITW & $--$ &
\textbf{\textcolor{blue}{\textsl{\datasetname}}}                            
& \textbf{10k+/10k+}  & ITW & $--$ \\
    \bottomrule
    \end{tabular}
\caption{
    Statistics on different human datasets for avatar creation. We only consider human datasets that include human appearances. \mcs: multi-camera system. \cs: 3D scanner. \ds: depth sensor. \imu: inertial measurement unit. \pa: professional actor. Our \datasetname is a \textit{large-scale} and \textit{in-the-wild} avatar dataset collected with our designed \textit{automatic collection} pipeline. 
  }
  
  \label{tab:previous-datasets}
\end{table*}

3D Human Avatar creation has extensive applications in VR/AR, film-making, metaverse, \etc, attracting significant attention recently. With the advent of neural radiance fields (NeRF)~\cite{nerf,AdvancesNerf}, recent research aims to recover 3D avatars from 2D observations, enabling the synthesis of novel images from arbitrary viewpoints and body poses~\cite{humannerf,neuralbody,animatable-nerf,gauhuman,nhp,mps-nerf,sherf}. 
For example, many works focus on reconstructing animatable human models from well-annotated and calibrated multi-view images and videos~\cite{neuralbody, animatable-nerf, animatable_sdf, NHR}, and have achieved photo-realistic results in laboratory datasets~\cite{neuralbody, renderpeople, PeopleSnapshot}. 
However, their performances are restricted by small-scale indoor human data, especially for generalizable creation.
This is down to the fact that current avatar creation datasets are mainly collected via well-designed laboratory systems, which are expensive and time-consuming to collect. 
In addition, there still exists domain gaps between laboratory and real-world scenarios. 
As shown in~\Cref{tab:previous-datasets}, existing human data collections mostly rely on annotations from advanced devices, such as well-calibrated multi-view cameras~\cite{thuman2,thuman3,MPI-INF-3DHP,DNA-Rendering}, depth sensors~\cite{humman,thuman,thuman2,h36m}, IMUs~\cite{3dpw}, or expensive scanners~\cite{renderpeople,Hi4D}, as well as specialized actors and lightstages~\cite{DynaCap,HUMBI,AIST,freeman,ENeRF}. 
These ideal conditions require high costs against scalability, exhibit limited representations of the real world, and are unavailable in in-the-wild scenarios (\ie, monocular web videos) or consumer applications.

To tackle this problem, recent efforts are attempting to collect in-the-wild monocular human data~\cite{TikTok,ivsnet, mirrorhuman} from the web. However, they still rely heavily on costly manual interventions, making it difficult to scale up. Therefore, their inadequate diversity fails to meet the requirements for dealing with in-the-wild challenges of 3D avatar creation. Given the shortage of real-world human data, scaling up avatar creation from real-world scenarios is worth exploring.

To this end, we propose a novel pipeline with filtering protocols to extract human movements automatically from the web. Specifically, we first streamline the annotation process with off-the-shelf state-of-the-art annotation methods~\cite{yolo,human-in-4d,easymocap,sam}, \eg, YOLO~\cite{yolo}, Segment Anything~\cite{sam}. 
We then propose a suite of assessment protocols for automatic filtering to retain only qualified video clips (\eg, without occlusions and annotated in high confidence).
We report our performance on the EMDB benchmark to investigate our pipeline's performance in real-world scenarios. With the help of the additional stage for aligning and smoothing, our pipeline surpasses state-of-the-art methods (up to $4\%$). 
We also demonstrate that the proposed filtering protocols can boost verification metrics among web video sources.

We implement the proposed annotating and filtering pipeline to collect real-world human samples from the web. Without costly sensors or lightstages, we curate \datasetname, a large-scale in-the-wild human avatar creation dataset extracted from YouTube, with $10,000+$ different real-world human subjects and scenes. \datasetname fills the gap in large-scale in-the-wild human data collection, offering at least $10\times$ richer human subjects than the existing human datasets.

To explore the potential of \datasetname, we investigate the scalability of existing generalizable human avatar creation methods~\cite{nhp,mps-nerf,sherf} with \datasetname, which suggests a significant improvement (up to $7\%$) in real-world scenarios. 
We will release the \datasetname dataset, providing the video IDs, frame IDs, extracting scripts, and annotations obtained from our pipeline. 
We hope that our dataset will push forward the development of 3D human avatar creation and other related topics, \eg, human mesh estimation (HPS)~\cite{hmr,pare,cliff,PyMAF,spin,vibe,simhmr,gvhmr}, 3D human avatar generation~\cite{chan2022efficient,hong2022eva3d,chen2023primdiffusion,hu2023humanliff,wang2023rodin,liu2023humangaussian}, 3D human interactions~\cite{HoloAssist,DBLP:conf/siggraph/ShuaiGFPSZB22}, and 3D relightable avatar~\cite{DBLP:conf/cvpr/0008P0MYSBZ24, RANA, Relighting4D}.
Overall, our main contributions can be summarized as follows:
\begin{itemize}
    \item We propose a novel annotating pipeline with filtering protocols to curate human movements from the web. Our pipeline surpasses state-of-the-art methods on the in-the-wild benchmark, and our filtering protocols can boost verification metrics on annotating web video sources.
    \item We curate \datasetname, a large-scale avatar dataset collected from in-the-wild videos with $10,000+$ human subjects, at least $10\times$ larger than previous datasets. The scale-up facilitates the creation of per-subject avatars and paves a new avenue for generalizable avatar reconstruction. 
    \item We illustrate the great potential of our pipeline and \datasetname, with exploratory experiments on supporting popular downstream 3D avatar creation applications. Further data scale-up for large-scale model training will be unlocked. The open code and data could provide important insights for future avatar creation and relevant tasks. 
\end{itemize}

%% file: sections/2_related_work.tex
\section{Related Work}
\label{sec:related_work}
\noindent\textbf{SMPL Annotation in the Wild.}
Parametric models (\eg, SMPL~\cite{SMPL}, SMPLX~\cite{SMPLX}) encode coarse human surfaces with pose and shape parameters for 3D avatar creation. For example, end-to-end methods~\cite{hmr, pare, PyMAF, cliff, simhmr} estimate the SMPL parameters efficiently from single in-the-wild images, but they are not robust with complex scenes. Subsequent works~\cite{spin, eft} refine end-to-end estimation results via fitting SMPL to 2D annotations in the loop. Recent works further consider imposing temporal consistency~\cite{easymocap,romp,vibe} to smooth SMPL estimations across in-the-wild videos. Yet, it is far from achieving a comprehensive annotating and filtering streamline for real-world video sources.

\noindent\textbf{3D Avatar Creation Datasets.} \Cref{tab:previous-datasets} presents a system overview of previous avatar creation datasets. 
Previous datasets mainly obtain high-quality human mask and SMPL annotations with ideal laboratory systems, \eg, well-calibrated multi-view cameras~\cite{neuralbody,humman,PKU-DyMVHumans,thuman2,h36m,thuman3,thuman4,thuman5,Hi4D,DNA-Rendering,DynaCap,multihuman,UltraStage,HumanRF,HUMBI,NHR,AIST,MPI-INF-3DHP,ENeRF,3dpw}, depth sensors~\cite{humman,thuman,thuman2,h36m,thuman3,DNA-Rendering}, IMUs~\cite{3dpw}, or expensive scanners~\cite{renderpeople,Hi4D}, as well as specialized actors and lightstages~\cite{neuralbody,renderpeople,humman,thuman,PKU-DyMVHumans,thuman2,h36m,thuman3,thuman4,thuman5,Hi4D,DNA-Rendering,DynaCap,multihuman,UltraStage,HumanRF,HUMBI,NHR,AIST,MPI-INF-3DHP,ENeRF,3dpw}, depth sensors~\cite{humman,thuman,thuman2,h36m,thuman3,DNA-Rendering}. 
Unfortunately, these optimal conditions are typically high-cost against scaling up and limited in scenarios compared to the real world. 
Towards in-the-wild avatar creation, previous efforts also attempt to collect avatar data from web human movement videos~\cite{TikTok,ivsnet}.
However, they rely heavily on costly manual interventions (\eg, pre-filtering or mask extraction based on ~\cite{backgroundmattingv2}) and still fail to scale up. And they either show little viewpoint change~\cite{TikTok} or not public released~\cite{ivsnet}. Therefore, there are still significant demands for a public large-scale in-the-wild dataset for avatar creation.

\noindent\textbf{3D Avatar Creation Methods.} Existing works on avatar creation learn coarse human surfaces with parametric mesh models~\cite{SCAPE, SMPL, SMPLX, GHUM, EmbodiedHands}, but these explicit meshes cannot express detailed geometry or appearance. Pifu-based methods~\cite{pifu, pifuhd, geo-pifu, DBLP:journals/pami/LiYZL22, PINA} represent the human body with pixel-aligned functions and require high-quality synthetic data, but fail to generalize to real-world scenarios due to the domain gap. NeRF-based approaches~\cite{neuralbody, animatable-nerf, humannerf, animatable_sdf, instantNVR, DBLP:conf/eccv/LiXSHOL20, DBLP:conf/cvpr/BozicPZTDN21, S3} learn implicit human representation from high-quality videos, and achieve photo-realistic rendering in the laboratory benchmarks. In addition, generalizable models~\cite{nhp, mps-nerf, sherf} further simplify the data demand to a single image. Nonetheless, these methods still rely on accurate annotations, which only exist in ideal laboratory environments. Recent attempts~\cite{neuman, Vid2Avatar} decompose avatars and scenes for construction at once, demanding accurate global alignments between human bodies and backgrounds that are not available in real-world videos. 
Web-scale in-the-wild video data is crucial for the next phase of 3D avatar development.

%% file: sections/3_method.tex
\section{Methodology}
\label{sec:dataset}
Our goal is to build a large-scale in-the-wild dataset for human avatar creation. 
To this end, we collect abundant real-world human videos from YouTube and annotate corresponding labels. 
Without time-consuming and high-cost manual filtering and annotation~\cite{ivsnet}, we design an efficient pipeline to filter video candidates and obtain high-quality annotations. 
Specifically, we download human movement videos from the web with our automatic tools, as described in~\Cref{subsec:data-collection}. Then, we collect high-quality annotated data with several processing steps on downloaded web videos (\eg, annotating in~\Cref{subsec:data-processing} and filtering in~\Cref{subsec:data-filtering}). 
Finally, we evaluate our designs in detail in~\Cref{subsec:pipeline-evaluation}.

\subsection{Data Collection}

\begin{figure*}[!t]
  \centering
    \includegraphics[width=0.99\linewidth]{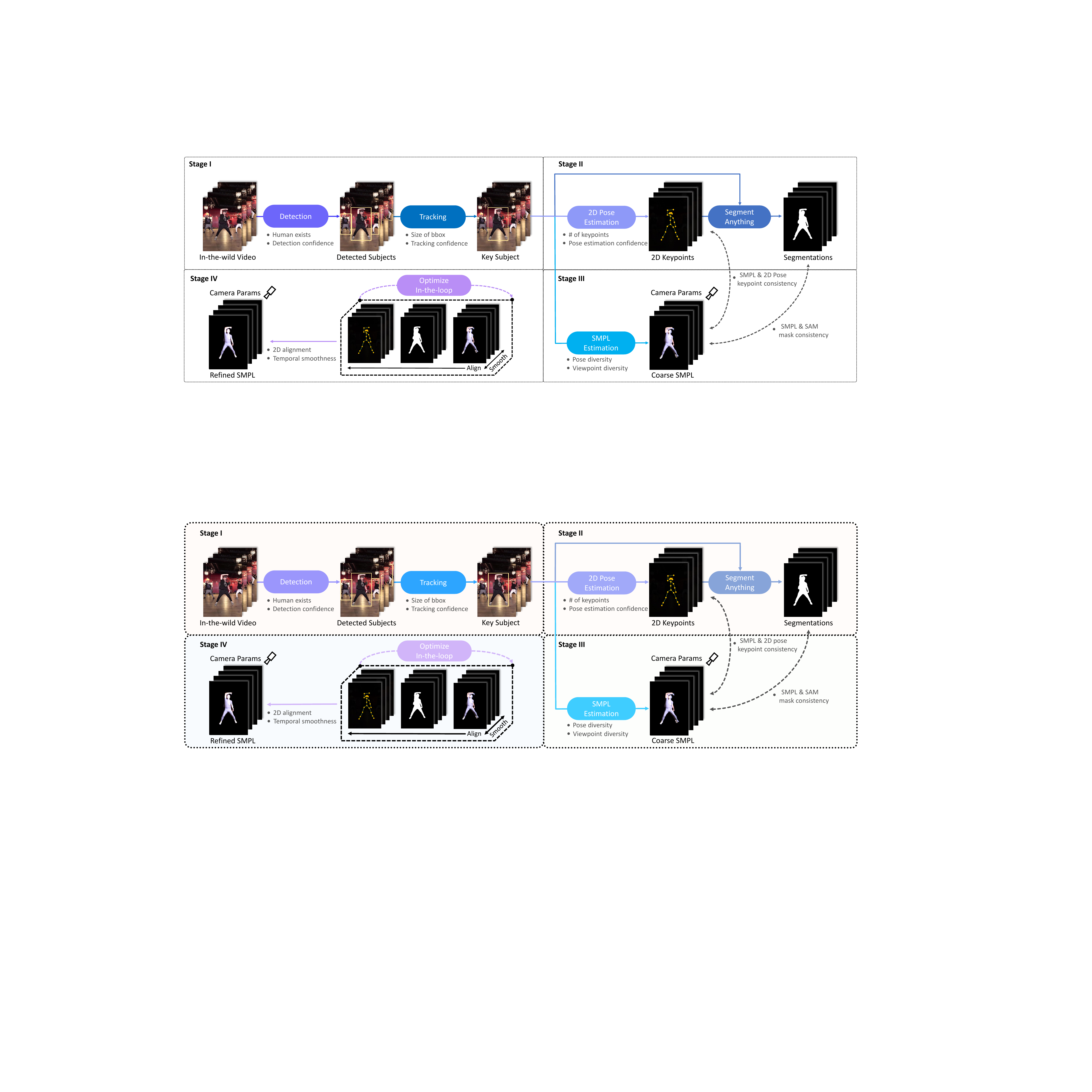}
    \caption{The four-stage data processing pipeline. We first obtain the bounding box of key subjects in videos in Stage I and extract human segmentation masks in Stage II. Then, the SMPL and camera parameters are coarsely estimated in Stage III and later refined in Stage IV.}
   \label{fig:pipeline}
   \vspace{-15pt}
\end{figure*}

\label{subsec:data-collection}
We follow two steps to collect human movement videos:
1) We first search for a large and diverse set of video candidates on YouTube. Note that video candidates could contain some noisy videos without human subjects; 
2) We then download the candidate videos with automatic tools~\cite{yt-dlp}, cut them into manageable clips, and filter out short video clips. Details can be found in~\Cref{sec:appendix-details-in-data-pipeline}.

\subsection{Data Annotating}
\label{subsec:data-processing}

To obtain high-quality annotations (\ie, SMPL, camera parameters, and human segmentation mask) for the collected human videos, we design the following four stages to annotate the in-the-wild human video clips automatically.

\noindent\textbf{Stage I: Human Bounding Box Detection and Tracking.}
Precise human bounding boxes are important for monocular SMPL estimation methods~\cite{pare, hmr, cliff, PyMAF} and human segmentation (using \textit{Segment Anything}~\cite{sam}). 
Hence, we first obtain the bounding box of human subjects with off-the-shelf state-of-the-art detection methods~\cite{yolo, detectron2}. 

\noindent\textbf{Stage II: Human Segmentation Mask Extraction.}
Foreground segmentation masks that distinguish human subjects from the background are important for 3D avatar creation. Previous methods~\cite{ivsnet, humannerf} adopt the \textit{background matting}~\cite{backgroundmattingv2} methods for segmentation. These methods, however, require manual intervention to select the collected videos with at least one still background image, which is time-consuming to prepare. 
To efficiently segment human video clips, we adopt state-of-the-art \textit{Segment Anything} (SAM)~\cite{sam}, which only requires bounding boxes and keypoints for each frame obtained in previous human detection, tracking, and 2D pose estimation steps (See~\Cref{fig:pipeline}). 

\noindent\textbf{Stage III \& IV: SMPL and Camera Estimation.}
We first estimate SMPL and camera parameters frame by frame, using state-of-the-art single-image-based human pose and shape estimation methods~\cite{pare, human-in-4d} in Stage III. 
However, these coarse parameters may ignore the temporal consistency of human motions. 
Inspired by previous SMPL annotation pipelines~\cite{spin, easymocap, eft}, we further refine SMPL and camera parameters in Stage IV, smoothing these parameters across the whole video clip sequence via gradient descent. 
We also incorporate the estimated 2D keypoints and SAM masks into a refinement loop, providing additional supervision for precise SMPL and camera parameter estimation. Qualitative comparisons between coarse and refined SMPL and camera parameters can be found in \Cref{fig:appendix-coarse-refined-comparison}.

\subsection{Data Filtering}
\label{subsec:data-filtering}
\begin{figure*}[!t]
  \centering
    \includegraphics[width=0.99\linewidth]{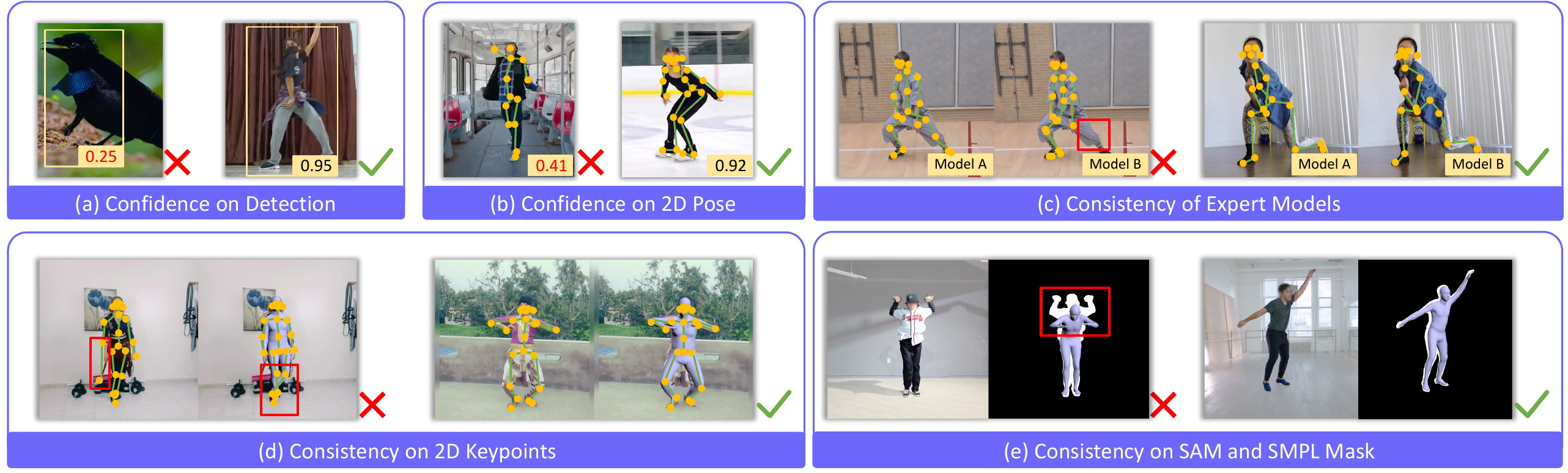}
   \vspace{-8pt}
    \caption{Visualizations of filtering protocols. We only retain video clips that (a) show high confidence in human detections; (b) obtain high average confidence in 2D pose estimations; (c) consistency annotated by different expert models; (d) consistency on keypoints between projected SMPL keypoints and 2D pose estimations; and (e) consistency on segmentation masks between Segment-Anything and SMPL.}
   \label{fig:filtering}
   \vspace{-12pt}
\end{figure*}

In contrast to high-quality human videos collected in well-designed laboratory systems, real-world videos are not always qualified for annotating or curating into datasets. To automatically filter out these unqualified videos (\eg, in severe occlusions or low resolution), we propose a suite of assessment protocols within the four-stage pipeline.

\noindent\textbf{Protocol I: Clear Body with Significant Movement.} 
The primary focus is to ensure the presence of a clear human body with no occlusion. To this end, we only retain in-the-wild videos with high confidence in both detection and 2D pose estimation, as illustrated in~\Cref{fig:filtering} (a) and (b). Then, we further reduce occlusion by only focusing on the key subject within each video clip. This is achieved by calculating the average size of the bounding box that encloses the tracked subject throughout the entire video. We also eliminate those trivial human movement videos, specifically those too brief or with insignificant viewpoint shifts and human movements.

\noindent\textbf{Protocol II: Ensemble of Annotating Experts.}
To ensure the quality of in-the-wild annotation, we adopt the ensemble of different state-of-the-art annotating model experts.
Specifically, we denote the model libraries for detection~\cite{yolo, detectron2}, 2D pose estimation~\cite{hrnet, dwpose, openpose}, and SMPL Estimation~\cite{pare, human-in-4d} as $\{\mathcal{D}, \mathcal{J}, \mathcal{S}\}$, and their predictions as $\{D_i, J_i, S_i\}$. 
We calculate the average $\{\mu(D_i), \mu(J_i), \mu(S_i)\}$ as the fused predictions, and the standard deviation $\{\sigma(D_i), \sigma(J_i), \sigma(S_i)\}$ as references for filtering. To reduce annotating errors, we only retain the videos that are annotated consistently among different state-of-the-art annotating experts (see~\Cref{fig:filtering} (c)). 

\noindent\textbf{Protocol III: Consistency on 2D Keypoints.} 
To augment the reliability of SMPL estimations, we implement an additional double-check of the consistency in 2D keypoints, between the monocular SMPL estimation and 2D pose estimation results. Specially, we project the 3D SMPL keypoints $J_{3D}(\mu(S_i))$ to 2D. Then, we compute the Percentage of Correct Keypoints (PCK)~\cite{PCK} between the projected $\mathbf{\Pi}(J_{3D}(\mu(S_i)))$ and those predicted from 2D pose estimation $\mu(J_i)$. Subsequently, we discard video clips that exhibit low average PCK values to ensure that only videos with high annotation confidence are retained for further curation.

\noindent\textbf{Protocol IV: Consistency in SMPL and SAM Masks.}
With inaccurate predictions, monocular SMPL estimation and SAM might have low overlaps in estimated human masks. 
We double-check their consistency by comparing SMPL projection masks with the SAM masks. 
Intuitively, the SMPL mask denotes the naked body, while the SAM mask contains the clothed body. 
Therefore, the SMPL mask should be mostly covered by the SAM mask. 
To guarantee high-quality annotations, we discard video clips whose SAM-masked region has small overlaps with the SMPL-masked region, as illustrated in~\Cref{fig:filtering} (e) (more examples in \Cref{fig:appendix-smpl-sam-comparison}).

\subsection{Analysis}
\label{subsec:pipeline-evaluation}
\noindent\textbf{Evaluation of Annotating Pipeline.}
To demonstrate the accuracy of our four-stage pipeline among in-the-wild videos, we first provide quantitative analysis on real-world EMDB~\cite{EMDB} benchmark in~\Cref{tab:emdb-benchmark}. Due to the additional 2D aligning and temporal smoothing in Stage IV, our annotations have more accurate alignment on extremities (\eg, feet, arms, and heads). Our pipeline surpasses existing methods, especially on the MPJPE~\cite{h36m} ($+3.27\%$) and PVE~\cite{XueCZYMM22} ($+4.16\%$), where global alignments are not taken into account.  Evaluation details and qualitative comparisons can be found in~\Cref{sec:appendix-details-in-pipeline-evaluation}. 

\begin{table}[!t]
  \centering
  \small
  \setlength{\tabcolsep}{11pt}

  \begin{tabular}{c|ccc}
    \toprule
     Method & PA-MPJPE$\downarrow$    & MPJPE$\downarrow$  & PVE$\downarrow$ \\ \midrule
     SPIN~\cite{spin}   
     & 87.1 & 140.3 & 174.9    \\
     VIBE~\cite{vibe}   
     & 81.4 & 125.9  & 146.8     \\
     PARE~\cite{pare}              
     & 72.2 & 113.9 & 133.2    \\
     TRACE~\cite{trace}   
     & 70.9 & 109.9  & 127.4     \\
     CLIFF~\cite{cliff}   
     & 68.1 & 103.3 & 128.0    \\
     TCMR~\cite{tcmr}   
     & 65.6 & 119.1  & 137.7     \\
     HybrIK~\cite{HybrIK}   
     & 65.6 & 103.0 & 122.2    \\
     HMR2.0~\cite{human-in-4d}                
     & 60.6 & 98.0 & 120.3     \\ 
     Ours                  
     & \textbf{59.9} & \textbf{94.9}  & \textbf{115.5}    \\
     \bottomrule
    \end{tabular}
    \vspace{-1pt}
    \caption{
    Performance comparisons on the public EMDB~\cite{EMDB} benchmark. Our proposed four-stage pipeline surpasses existing methods on the quality of SMPL annotations.
    }
  \label{tab:emdb-benchmark}
  \vspace{-20pt}
\end{table}

\noindent\textbf{Verification of Filtering Strategies.}
We further demonstrate the efficacy of our filtering protocols and validate the quality of our final dataset. 
Given that there is no ground truth (GT) for evaluation, we adopt double-checks for verification to assess the quality of annotations for in-the-wild data. 
Specifically, we evaluate the consistency between predicted annotations after applying our protocols. The detailed results are presented in~\Cref{tab:evaluation-protocols}, where PCK represents the consistency between the 2D pose estimation models and the SMPL models, while the SOIOU measures the proportion of the SMPL masks that lie outside the SAM masks. Comparing adjacent columns, we find our pipeline designs compelling: 1) Protocol I is a vital pre-processing, which brings a marginal improvement on PCK and SOIOU. It eliminates a substantial number of clips in poor quality (\eg, no human subjects or movements); 2) Protocol II also yields a significant enhancement on PCK ($10.1\%$), demonstrating the importance of adopting various annotating experts. 3) Protocol III \& IV step further towards consistency, raise PCK by $7.51\%$ and drop SOIOU by $0.094$ in total; and 4) Stage IV finally refines the SMPL annotations, resulting in PCK over $0.92$ and SOIOU lower than $0.03$. Notably, this final PCK deviates by only $1.7\%$ from the 3DPW dataset.

\begin{table}[!t]
  \centering
  \small
  \setlength{\tabcolsep}{5pt}

  \begin{tabular}{ccccc|cc|c}
    \toprule
     P.I & P.II  & P.III & P.IV & S.IV & PCK$\uparrow$ & SOIOU$\downarrow$ & \#Sub.\\ \midrule
     $    --    $ & $    --    $ & $    --    $ & $    --    $ & $    --    $ & 0.282 & 0.760 &  465801 \\
     $\checkmark$ & $    --    $ & $    --    $ & $    --    $ & $    --    $ & 0.762 & 0.214 &  43824 \\
     $\checkmark$ & $\checkmark$ & $    --    $ & $    --    $ & $    --    $ & 0.839 & 0.146 &  25392 \\
     $\checkmark$ & $\checkmark$ & $\checkmark$ & $    --    $ & $    --    $ & 0.882 & 0.129 &  12482 \\
     $\checkmark$ & $\checkmark$ & $\checkmark$ & $\checkmark$ & $    --    $ & 0.902 & 0.052 &  10647 \\
     $\checkmark$ & $\checkmark$ & $\checkmark$ & $\checkmark$ & $\checkmark$ & \textbf{0.921} & \textbf{0.028} &  10647 \\ 
     \midrule
     \multicolumn{5}{c|}{Ground Truth of 3DPW} & 0.937 & $--$ &  7  \\
     \bottomrule
    \end{tabular}
\vspace{-6pt}
\caption{
    Evaluation of the proposed pipeline on web videos. PCK (threshold $0.1$) represents the consistency between the 2D pose and SMPL estimation, while SOIOU measures the proportion of the SMPL mask outside the SAM mask. \#Sub. denotes the number of qualified video clips. P.X/S.X denotes the Protocol/Stage X. 
    }
  \label{tab:evaluation-protocols}
  \vspace{-15pt}
\end{table}

\begin{figure*}[!t]
  \centering
    \includegraphics[width=0.99\linewidth]{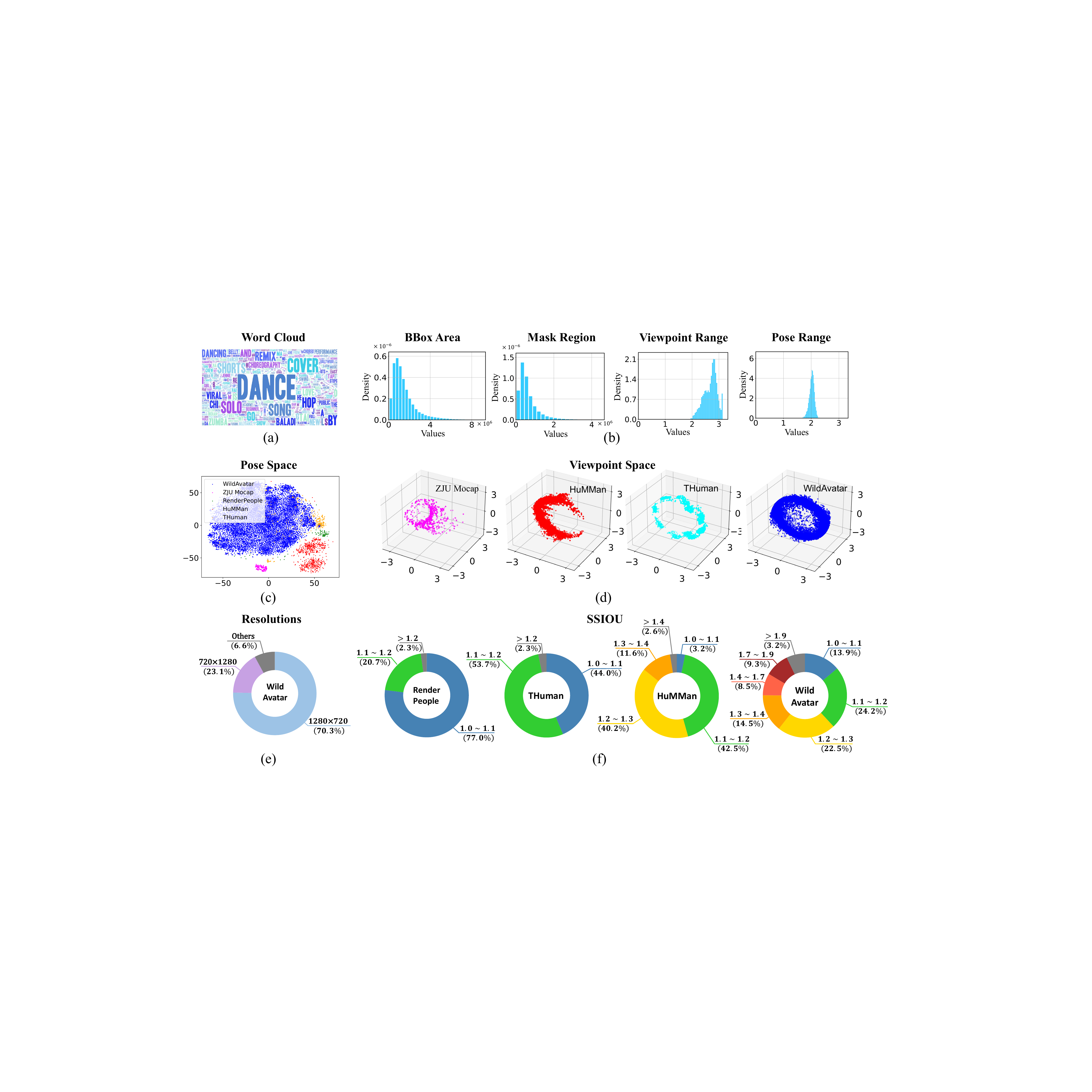}
    \vspace{-8pt}
    \caption{Data Analysis: (a) word cloud of the video titles in \datasetname, (b) histograms of annotations across video clips, here we count the bounding box and human mask region in pixels, and ``Range'' denotes the difference between the maximum and minimum values. (c) comparison of the body pose spaces with popular laboratory human datasets, (d) comparison of the viewpoints spaces with popular laboratory human datasets, (e) resolutions of videos in \datasetname, and (f) comparison with the previous dataset on the abundance of clothing. We introduce the SSIOU, the inverse IOU between SMPL~\cite{SMPL, SMPLX} masks and segmentation masks.}
   \label{fig:data-analysis}
   \vspace{-15pt}
\end{figure*}

%% file: sections/4_dataset.tex
\section{\datasetname}
\label{subsec:data-analysis}
Our collected \datasetname contains $10,647$ real-world human video clips. We randomly split \datasetname into training, validation, and test splits with $7k$, $1.5k$, and $1.5k$ subjects, respectively. Notably, although selected by our protocols, the remaining curated data is still much more diverse than in the laboratory systems (See~\Cref{subsec:data-analysis} for statistics and~\Cref{fig:qualitative-comparison} for unusual camera viewpoints and body poses). We show data statistics of our \datasetname in~\Cref{fig:data-analysis} and describe the details of pose, viewpoint, and clothing distribution as follows. 

\noindent\textbf{Poses: Various vs. Trivial.}
In~\Cref{fig:data-analysis} (c), we compare the pose distribution differences among \datasetname and previous human datasets. 
Specifically, we visualize the top-2 components of body poses with t-SNE~\cite{tsne} dimension reduction. 
The pose distribution comparison shows that our \datasetname contains more diverse poses than previous laboratory datasets, as laboratory datasets only contain several designed motion sequences.
This statistic illustrates the importance of our large-scale in-the-wild \datasetname when applying the existing models to real-world scenarios. 

\noindent\textbf{Viewpoints: Free vs. Fixed.}
We compare viewpoint distribution among previous human datasets~\cite{neuralbody, humman, thuman} and \datasetname. Specifically, we visualize the 3D rotations of observed cameras relative to human subjects.
As suggested in~\Cref{tab:previous-datasets}, previous laboratory datasets are mainly collected in indoor lightstages, where the RGB(D) cameras' positions are fixed for whole datasets.
In real-world scenarios, however, the viewpoints of humans are arbitrary. 
As shown in~\Cref{fig:data-analysis} (d), the viewpoint distributions in previous datasets are sparse or unbalanced, while our \datasetname covers a more dense viewpoint distribution.

\noindent\textbf{Cloth: Diversity vs. Unitary.}
Previous laboratory human datasets mainly contain unitary tight clothes, while complex real-world appearances (\eg, different hairstyles) and diverse loose clothes are seldom involved. Our
\datasetname involves diverse tight and loose clothes in the real world. 
To quantitatively measure the clothing diversity between real-world and laboratory data, we define a metric (SSIOU) that calculates the inverse IOU between SMPL~\cite{SMPL, SMPLX} projected masks and human segmentation masks among different datasets in~\Cref{fig:data-analysis} (f). 
For humans with tight clothes, the measured SSIOU is close to $1.0$ (blue in~\Cref{fig:data-analysis} (f)), indicating their SMPL masks are similar to human segmentation masks. Correspondingly, a larger SSIOU (red in~\Cref{fig:data-analysis} (f)) reveals that human clothes are looser, with segmentation masks being larger than SMPL masks. Examples of various SSIOU can be found in~\Cref{fig:ssiou}.
As shown in~\Cref{fig:data-analysis} (f), more than $30\%$ subjects in \datasetname have  SSIOU values over $1.4$, suggesting that \datasetname covers various types of clothes.
This analysis validates that \datasetname more satisfies data distributions in real-world scenarios.

%% file: sections/5_experiments.tex
\section{Experiments}
\label{sec:experiments}

\begin{table}[!t]
  \centering
  \small
  \setlength{\tabcolsep}{3pt}
  \begin{tabular}{c|ccc|ccc}
    \toprule
     \multirow{2}{*}{\textbf{Method}}  & \multicolumn{3}{c|}{\textbf{HMR2.0}} & \multicolumn{3}{c}{\textbf{Ours}} \\
     & PSNR$\uparrow$ & SSIM$\uparrow$    & LPIPS$\downarrow$  & PSNR$\uparrow$ & SSIM$\uparrow$    & LPIPS$\downarrow$ \\ \midrule
     NHR~\cite{NHR}   
     & 18.23      & 89.7      & 10.3   & 18.93      & 91.6      & 9.9     \\
     NB~\cite{neuralbody}              
     & 16.17      & 84.2      & 12.6   & 16.74      & 86.2      & 12.1    \\
     AN~\cite{animatable-nerf}   
     & 19.02      & 89.1      & 11.4   & 19.72      & 90.5      & 10.9    \\
     AS~\cite{animatable_sdf}   
     & 18.72      & 90.9      & 11.1   & 19.20      & 92.3      & 10.7    \\
     HN~\cite{humannerf}                
     & 22.52      & 86.3      & 15.2   & 23.12      & 88.0      & 14.6    \\
     IN~\cite{instantNVR}   
     & 24.39      & 92.7      &  8.2   & 25.28      & 94.3      &  7.7    \\
     GH~\cite{gauhuman}                  
     & 24.73      & 93.8      & 6.3    & 25.89      & 95.7      & 5.7     \\
     \bottomrule
    \end{tabular}
    \caption{
    Quantitative comparisons on \datasetname. We report the quality of novel pose synthesis of popular methods on the state-of-the-art HMR2.0 annotations and our annotations, revealing the advantages of our pipeline towards in-the-wild avatar creation.
    }
    \vspace{-5pt}
  \label{tab:persubject-youtube-benchmark}
\end{table}

\begin{figure}[!ht]
  \centering
    \includegraphics[width=0.99\linewidth]{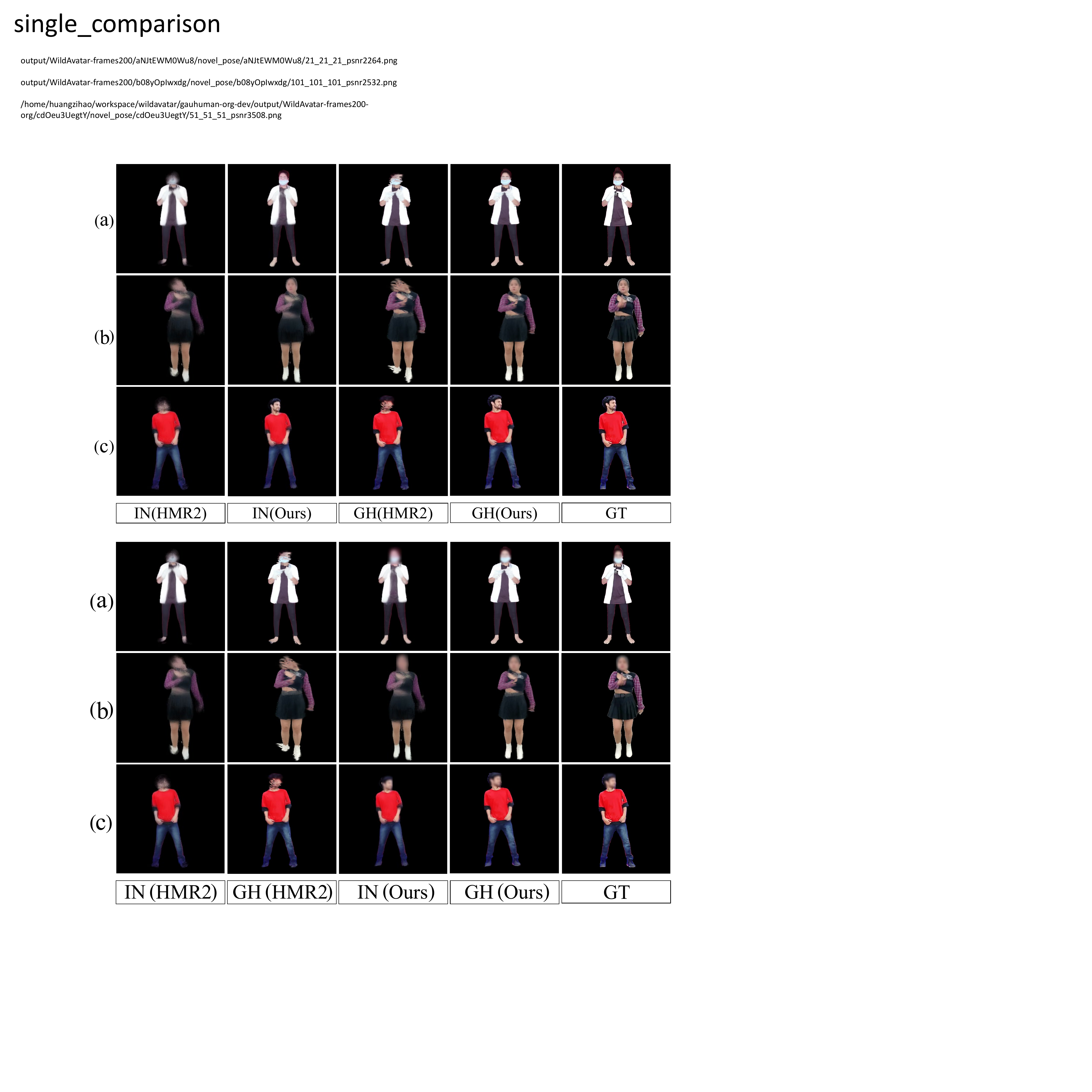}
    \caption{Qualitative comparisons of avatars created with our annotations and the state-of-the-art HMR2.0. IN and GH denote InstantNVR and GauHuman, respectively. More accurate avatars can be created with our annotations. Human faces are blurred.
    }
    \vspace{-15pt}
    \label{fig:persubject-compare}
\end{figure}

We conduct exploratory experiments on our \datasetname with the following commonly used metrics (details in~\Cref{subsec:metrics}) on 3D avatar creation. 
~\Cref{subsec:persubject-experiments} illustrates the real-world avatars from our \datasetname, created by per-subject avatar creation methods. It also illustrates the improvement of our annotating pipeline, by comparing annotations from our pipeline or the state-of-the-art HMR2.0~\cite{human-in-4d}.
~\Cref{subsec:generalizable-experiments} further demonstrates the potential of generalizable avatar creation methods, when provided our large-scale \datasetname. 

\begin{table*}[!t]
  \small
  \centering
  \setlength{\tabcolsep}{7pt}
  \begin{tabular}{c|c|cccc|ccc|cc}
    \toprule
     \multirow{2}{*}{\textbf{Method}} & \multicolumn{5}{c|}{\textbf{Training Set}} &  \multicolumn{3}{c|}{\textbf{\datasetname Test}} & \multicolumn{2}{c}{\textbf{Novel Lab Test}} \\
    & \textcolor{blue}{\textbf{WA}} & RP & THU & HM & ZJU & PSNR(NP) & SSIM(NP) & LPIPS(NP) & PSNR(NV) & PSNR(NP)\\ \midrule
    
     \multirow{4}{*}{NHP~\cite{nhp}} 
    &          &          &\checkmark&\checkmark&\checkmark&
    17.39&62.6&33.2&20.05&19.83 \\
    &\textcolor{blue}{\checkmark}&          &\checkmark&\checkmark&\checkmark&
    \best18.12&\best66.3&\best31.4&\best20.49&\best20.24 \\
    &          &\checkmark&          &\checkmark&\checkmark&
    17.35&62.2&33.4&20.84&20.81 \\
    &\textcolor{blue}{\checkmark}&\checkmark&          &\checkmark&\checkmark&
    \best18.07&\best66.9&\best31.2&\best21.20&\best21.27 \\ \midrule
     \multirow{4}{*}{MPS-NeRF~\cite{mps-nerf}} 
    &          &\checkmark&\checkmark&          &\checkmark
    &18.21&74.2&24.0&16.99&17.03 \\
    &\textcolor{blue}{\checkmark}&\checkmark&\checkmark&          &\checkmark&
    \best18.79&\best76.8&\best22.2&\best18.48&\best18.77 \\
    &          &\checkmark&\checkmark&\checkmark&          &
    18.20&74.3&23.8&20.17&20.03\\
    &\textcolor{blue}{\checkmark}&\checkmark&\checkmark&\checkmark&          &
    \best19.11&\best77.2&\best22.1&\best21.19&\best21.48\\ \midrule
     \multirow{4}{*}{SHERF~\cite{sherf}}
    &          &          &\checkmark&\checkmark&\checkmark&
    18.39&78.1&20.2&20.67&20.10\\
    &\textcolor{blue}{\checkmark}&          &\checkmark&\checkmark&\checkmark&
    \best19.43&\best80.5&\best19.3&\best20.96&\best20.49\\
    &          &\checkmark&\checkmark&          &\checkmark&
    18.31&77.9&20.4&19.07&18.99\\
    &\textcolor{blue}{\checkmark}&\checkmark&\checkmark&          &\checkmark&
    \best19.50&\best80.7&\best19.2&\best19.57&\best19.64\\ \midrule
     \multirow{4}{*}{Generalizable GS}
    &          &\checkmark&          &\checkmark&\checkmark&
    17.76&81.1&22.8&20.31&20.03\\
    &\textcolor{blue}{\checkmark}&\checkmark&          &\checkmark&\checkmark&
    \best18.43&\best81.5&\best22.3&\best21.73&\best21.65\\
    &          &\checkmark&\checkmark&\checkmark&          &
    17.73&81.2&22.7&21.47&21.78\\
    &\textcolor{blue}{\checkmark}&\checkmark&\checkmark&\checkmark&          &
    \best18.39&\best81.4&\best22.4&\best22.46&\best22.02\\ 
     \bottomrule
    \end{tabular}
\caption{
    Generalizability comparisons on challenging \datasetname and laboratory benchmarks. We report the results of previous generalizable avatar creation methods on the cross-domain setting, offering a clear perspective across different domains. NP/NV denotes Novel Pose/View. RP/THU/HM/ZJU/\textcolor{blue}{\textbf{WA}} are short for RenderPeople~\cite{renderpeople} / THuman~\cite{thuman} / HuMMan~\cite{humman} / ZJU Mocap~\cite{neuralbody} / \textcolor{blue}{\textbf{\datasetname (Ours)}} dataset. Novel Lab Test refers to the test split of laboratory datasets not included in the training. 
    }
    \vspace{5pt}
  \label{tab:cross-benchmark}
\end{table*}

\subsection{Evaluation Metrics}
\label{subsec:metrics}
To quantitatively evaluate the quality of created avatars, we apply three commonly used metrics: peak signal-to-noise ratio (PSNR)~\cite{psnr}, structural similarity index (SSIM)~\cite{ssim}, and Learned Perceptual Image Patch Similarity (LPIPS)~\cite{lpips}. 
Following previous literature~\cite{mps-nerf, sherf}, we compute whole-image metrics for per-subject reconstruction methods, while reporting the metrics based on projected 3D human bounding box areas for generalizable methods.

\subsection{Per-Subject Avatar Creation}
\label{subsec:persubject-experiments}

\noindent\textbf{Setup.}
To explore the quality of \datasetname and investigate the reconstruction performance on monocular in-the-wild videos with our annotations. We evaluate popular per-subject avatar creation methods on our \datasetname dataset. We consider a total of 7 baselines: Neural Human Rendering (\underline{NHR})~\cite{NHR}, NeuralBody (\underline{NB})~\cite{neuralbody}, Animatable NeRF (\underline{AN})~\cite{animatable-nerf}, Animatable SDF (\underline{AS})~\cite{animatable_sdf}, HumanNeRF (\underline{HN})~\cite{humannerf}, InstantNVR (\underline{IN})~\cite{instantNVR}, and GauHuman (\underline{GH})~\cite{gauhuman}. Considering the calculating and time cost, it is impractical and inefficient to train all subjects and scenes in \datasetname. Instead, we manually select $133$ representative human subjects from \datasetname for exploring. For each subject, we randomly choose $100$ frames for training and $100$ for testing. We follow the default model settings (learning rate, batch size, \etc) in their official implementations and report average metrics of these baselines in \Cref{tab:persubject-youtube-benchmark}.

\noindent\textbf{Quantitative and Qualitative Analysis.}
As shown in~\Cref{tab:persubject-youtube-benchmark}, more accurate avatars can be created with our annotations. Compared to the annotations from the state-of-the-art HMR2.0~\cite{human-in-4d}, our annotations improve PSNR ($+3.51\%$), SSIM ($+1.91\%$), and LPIPS ($-5.06\%$) on average. We further present qualitative comparisons in~\Cref{fig:persubject-compare}. Avatars from our annotations have more accurate extremities (\eg, feet, arms, and heads), which are challenging to predict and align from single images. This improvement may prove the benefit of Stage IV (\eg, 2D alignment and temporal smoothness).

\subsection{Generalizable Avatar Creation}
\label{subsec:generalizable-experiments}

\begin{figure*}[!t]
  \centering
    \includegraphics[width=0.99\linewidth]{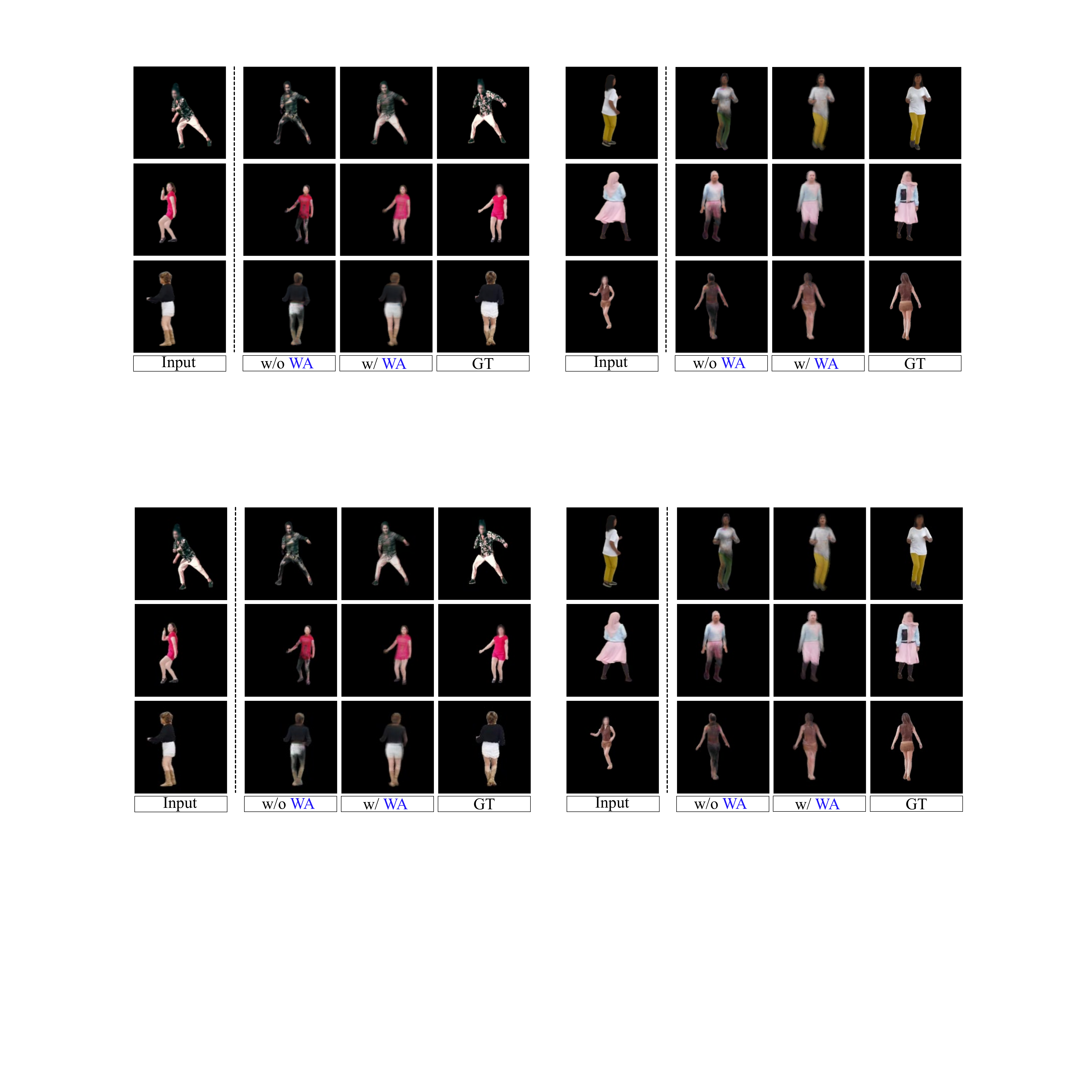}
  \vspace{-4pt}
    \caption{Qualitative comparisons on the state-of-the-art generalizable avatar creation method~\cite{sherf}.
    ``w/ WA'' or ``w/o WA'' denotes training with or without \datasetname, respectively. Human faces are blurred to protect privacy.
    }
    \label{fig:generalizable-compare}
\end{figure*}

\noindent\textbf{Setup.}
We evaluate four state-of-the-art generalizable baselines on our \datasetname, including 1) Neural Human Performer (\underline{NHP})~\cite{nhp}, 2) \underline{MPS-NeRF}~\cite{mps-nerf}, 3) \underline{SHERF}~\cite{sherf}, and 4) Generalizable 3D Gaussian Splatting (\underline{Generalizable GS}). Specifically, the Generalizable GS is adapted from GauHuman~\cite{gauhuman}, with the Gaussians' attributes decoded from the input features, following the approaches in ~\cite{pixelsplat,mvsgaussian}. To evaluate the overall generalizability among various scenarios, we also report results on four laboratory datasets, including 1) ZJU-Mocap~\cite{neuralbody}, 2) RenderPeople~\cite{renderpeople}, 3) HuMMan~\cite{humman}, and 4) THuman~\cite{thuman}. 
To avoid overfitting to a small indoor dataset, we adopt the cross-domain testing strategy, \eg, mask out the ZJU training set when testing on the ZJU.

\noindent\textbf{Quantitative and Qualitative Analysis.}
There are three main observations in~\cref{tab:cross-benchmark}. First, large-scale \datasetname is essential for improving generalizability towards in-the-wild scenarios. Comparing odd and even rows, the \datasetname training set brings considerable improvements in real-world scenarios. On the \datasetname test set, it improves PSNR, SSIM, and LPIPS by 4.52\%, 3.53\%, and 5.06\% on average, respectively. Notably, this is fairly evident as over $60\%$ of pixels are background. This improvement suggests that large-scale in-the-wild data is essential for the generalizable avatar creation methods. As shown in~\Cref{fig:generalizable-compare}, models without (w/o) training on \datasetname, tend to predict dark-colored artifacts on the lower body. Models with (w/) training on \datasetname can provide more realistic appearances and geometries. 
Second, large-scale in-the-wild data is also beneficial for laboratory benchmarks. However, there are significant domain gaps, and the annotation of in-the-wild data is not as accurate as in the laboratory. By introducing \datasetname dataset, we obtain an improvement of 4.17\% and 4.54\% PSNR on novel view and pose synthesis, respectively. 
Third, although 3D Gaussian Splatting (3DGS) performs well in per-subject human avatar creation, it shows slightly worse than NeRF in this generalizable setting. 
Our insight is that 3DGS highly relies on the point cloud initialization~\cite{gauhuman, gaussianpro, mvsgaussian}. Unfortunately, due to the depth ambiguity, it is challenging to infer accurate point cloud initialization from a single image.

%% file: sections/6_conclusion.tex
\section{Conclusion}
\vspace{-5pt}
\label{sec:conclusion}
This work introduces an automatic pipeline with filtering protocols for collecting in-the-wild human annotations from the web. From this pipeline, we curate \datasetname, a large-scale in-the-wild human avatar creation dataset from YouTube, with $10,000+$ various human subjects and scenes. Compared with traditional avatar creation datasets, our \datasetname consists of at least $10\times$ more subjects or scenes. We demonstrate the effectiveness of the proposed pipeline and illuminate the potential of \datasetname on avatar creation tasks with data. We hope our pipeline and dataset will shed insights on following in-the-wild human avatar creation and benefit 3D/4D human content generation works.

\noindent\textbf{Acknowledgement.} 
This study is supported under the RIE2020 Industry Alignment Fund – Industry Collaboration Projects (IAF-ICP) Funding Initiative, as well as cash and in-kind contribution from the industry partner(s).

%% file: sections/7_appendix.tex
\setcounter{figure}{0}
\renewcommand{\thefigure}{{\Alph{figure}}}
\renewcommand{\thesection}{{\Alph{section}}}

\begin{figure*}[!h]
  \centering    \includegraphics[width=0.99\linewidth]{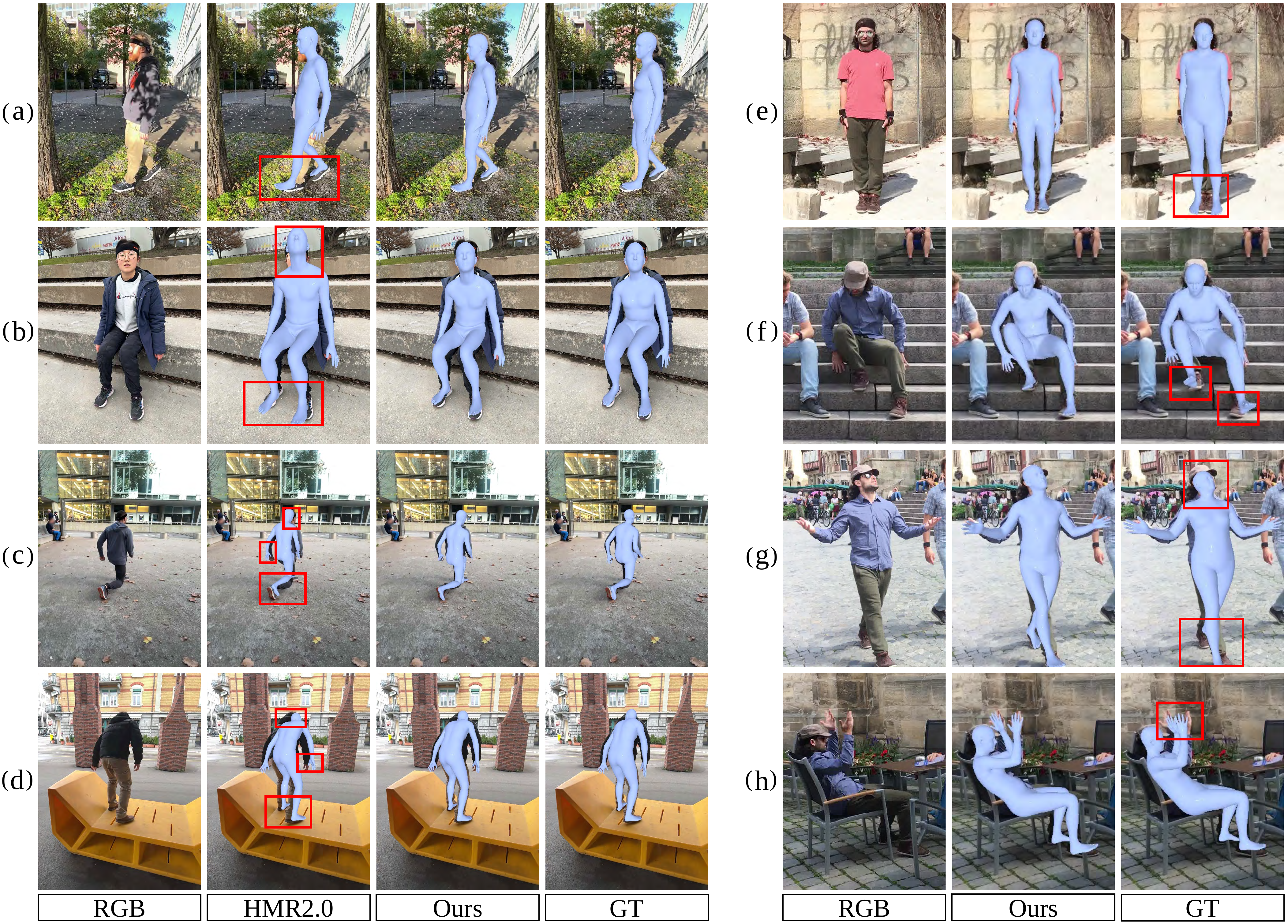}
    \caption{Qualitative comparison of our four-stage pipeline and the state-of-the-art HMR2.0~\cite{human-in-4d}. Our pipeline can adapt to complex environmental scenes and output reasonable results in uncommon scenarios.}
    \label{fig:qualitative-comparison}
\end{figure*}

\section{Details in Data Collection}
 \noindent\textbf{Downloading Video Candidates.} 
Our first goal is to collect videos from the web with human motions. To cover a wide range of in-the-wild human-central activities, we start from a label pool of human motion datasets~\cite{motionx}. Based on these human motion labels (\eg, fishing and playing tennis), we download over $100k+$ video candidates from YouTube API. 

\noindent\textbf{Per-filtering Video Clips.}
Some collected video candidates could not meet the high-quality human avatar creation requirement. 
For example, human bodies may not exist at all (\eg, blank preview, severe occlusion) or frequently change across scenes (\eg, montage) in some subsections of these videos. 
To exclude such unqualified subsections, we utilize SceneDetect~\cite{scenedetect} to cut video candidates into clips and eliminate those with insufficient length (less than $2$ seconds). Subsequently, we apply human detection models with low FPS to these clips to efficiently filter out those without human subjects at minimal cost.
After filtering video candidates, we obtain $460k+$ video clip candidates for further processing.

\section{Details in Data Pipeline}
\label{sec:appendix-details-in-data-pipeline}
In addition to the main paper, we provide more details on the data processing pipeline and filtering protocols.

\noindent\textbf{Stage I: Human Bounding Box Detection and Tracking.}
We first obtain the bounding box of human subjects with off-the-shelf state-of-the-art detection methods (\eg, Yolo~\cite{yolo} and Detectron2~\cite{detectron2}). We only keep the video clip with at least one ``person'' instance with its detection threshold over $0.8$ on all detection models. The tracking step is finding the largest IOU overlay of bounding boxes among frames. We discard low-resolution human subjects whose bounding box areas are lower than $64\times64$. To ensure the richness of the dataset, we only keep one ``key subject'' for each video, as clips from the same video may probably share the same key subject.

\noindent\textbf{Stage II: Human Segmentation Mask Extraction.}
We first obtain the 2D keypoints $J_{2D}$ for human subjects using the popular HRNet~\cite{hrnet} and DWPose~\cite{dwpose}. 
Given the 2D keypoint annotations, we can also discard over part-occluded subjects. In particular, we only keep the subject with the average confidence of 2D keypoints over $0.65$. For segmentation, we feed the 2D bounding box and the 2D keypoints into the \textit{sam\_vit\_h} sub-model to extract the foreground mask. 

\begin{figure*}[!t]
  \centering
    \includegraphics[width=0.99\linewidth]{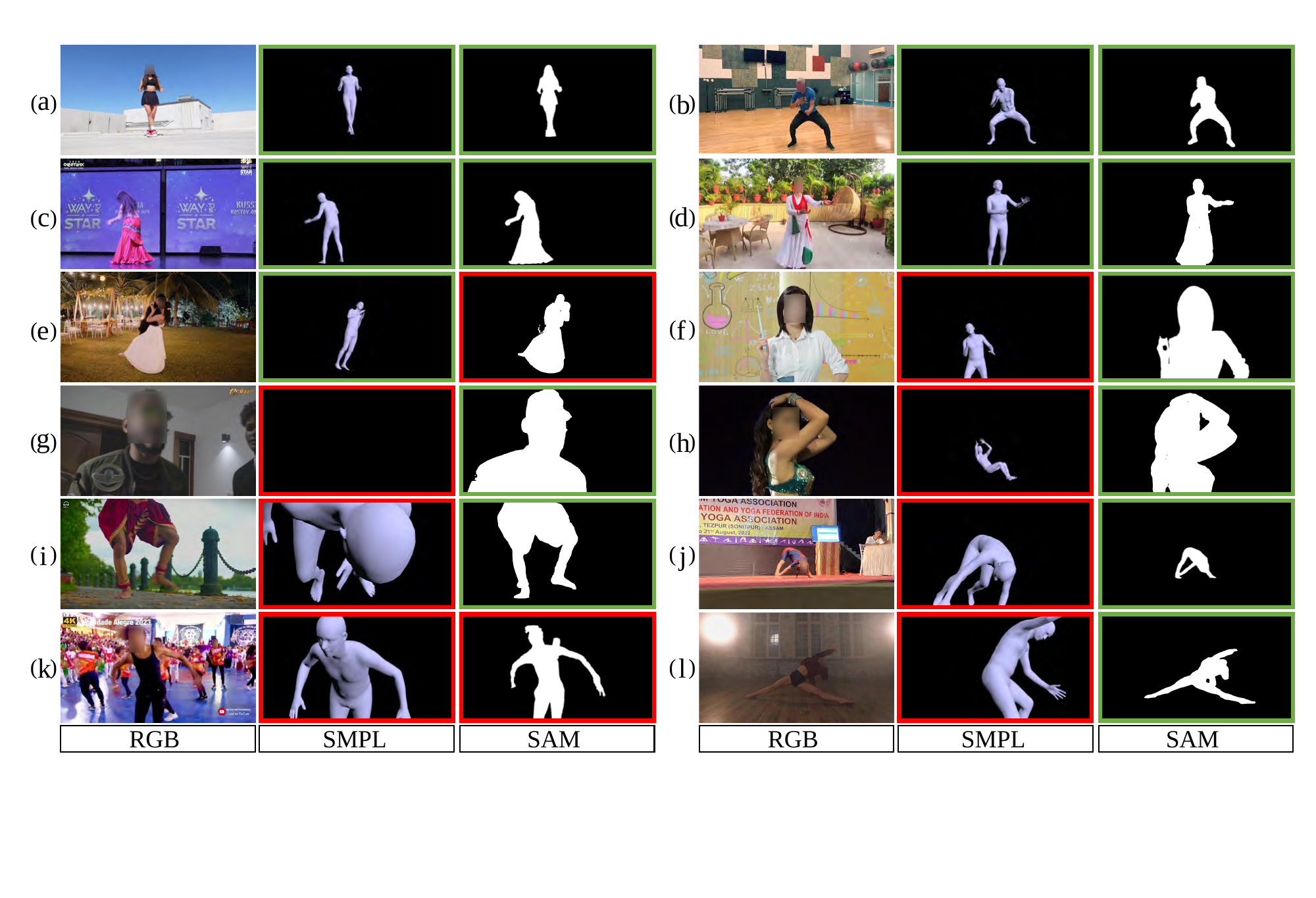}
    \caption{SMPL and SAM consistency. The green/red borders denote good/bad outputs, respectively. Note that the SMPL annotations are coarse results from Stage III, which are later refined in Stage IV. 
    }
    \label{fig:appendix-smpl-sam-comparison}
\end{figure*}

\noindent\textbf{Stage III: Coarse SMPL and Camera Estimation.}
We first estimate SMPL and camera parameters frame by frame, using state-of-the-art single-image-based human pose and shape (HPS) estimation methods~\cite{pare, human-in-4d}. To perform better in complex scenes in the wild, we adapt the model pre-trained on the in-the-wild 3DPW dataset~\cite{3dpw}. The HPS models infer human body pose/shape parameters ($\theta$/$\beta$) and the global camera parameters (rotation matrix $R$ and the 3D offset $T$). 
To retain the remaining video clips with considerable viewpoint shifts and human movements, we discard the clips with viewpoint angle changes lower than $\frac\pi4~rad$. We also automatically select the most non-trivial $N=20$ frames, which keeps the pose and viewpoint diversity to the greatest extent possible. 
As mentioned in the main paper, we double-check the consistency of the SAM and SMPL masks. Intuitively, the SMPL mask denotes the naked body, while the SAM mask contains the clothed body. Therefore, the SMPL mask should be mostly covered by the SAM mask (See~\Cref{fig:appendix-smpl-sam-comparison} (a) ~\textasciitilde~ (d)). We discard the subjects whose SAM masks are over $3\times$ larger than their SMPL masks (See~\Cref{fig:appendix-smpl-sam-comparison} (e) ~\textasciitilde~ (h)). We also discard the subjects whose $10\%$ SMPL mask pixels from main bodies are not covered by the SAM mask (See~\Cref{fig:appendix-smpl-sam-comparison} (i) ~\textasciitilde~ (l)). Similarly, we double-check the consistency of the 2D keypoints from 2D pose and SMPL estimations and discard the clips with the averaged PCK less than $0.85$.

\noindent\textbf{Stage IV: Refining SMPL and Camera In-the-loop.}
We refine the coarse SMPL parameters ($\theta, \beta$) and camera parameters ($R, T$) obtained in Stage I for high-quality annotations. 
To achieve temporally smooth results, we regularize the differences in parameters between adjacent frames, which is given by
\begin{equation}
\begin{aligned}
    \mathcal{L}^{s}_{\theta} &= \sum_{i=1}^{N-1}\left\|\theta^i - \theta^{i+1}\right\|_2, \\
    \mathcal{L}^{s}_{R} &= \sum_{i=1}^{N-1}\left\|R^i - R^{i+1}\right\|_2, \\
    \mathcal{L}^{s}_{T} &= \sum_{i=1}^{N-1}\left\|T^i - T^{i+1}\right\|_2, \\ 
    \mathcal{L}^{s}_{2D} &= \sum_{i=1}^{N-1}\left\|\Pi(J_{3D}(\theta^i, \beta^i); R^i, T^i) \right. \\
    &\quad \left. - \Pi(J_{3D}(\theta^{i+1}, \beta^{i+1}); R^{i+1}, T^{i+1})\right\|_2,
\end{aligned}
\end{equation}

where $J_{3D}(\theta,\beta)$ infers the 3D keypoints of the human body, and the $\Pi$ denotes the 2D projection, and $i$ denotes the $i_{th}$ frame of the input video.
Notice that the body shapes ($\beta$) are treated as constants across the input video.

In addition, we align the human body parameters to the 2D keypoints and SAM masks, which are given by
\begin{equation}
\begin{aligned}
\mathcal{L}_{2D} &= \sum_{i=1}^{N}\left\|\Pi(J_{3D}(\theta^{i},\beta^{i}))-J^{i}_{2D}\right\|_{2}, \\
    \mathcal{L}_{\mathrm{mask}} &=\sum_{i=1}^{N}(M_{rd}(\theta^i,\beta^i)), M^i),
\end{aligned}
\end{equation}
where $M_{rd}(\theta,\beta)$ denotes the rendered mask related to the SMPL parameters. The $M^i$ denotes the foreground of the $i$ frame. 
We also regularize $\theta$ to avoid out-of-domain poses~\cite{smplify}, using the Gaussian Mixture Model (GMM) prior~\cite{SMPLX}
\begin{equation}
    \mathcal{L}_{\mathrm{prior}}=\sum_{i=1}^{N}\left\|GMM(\theta^i)\right\|_2.
\end{equation}

\begin{figure*}[!t]
  \centering
    \includegraphics[width=0.99\linewidth]{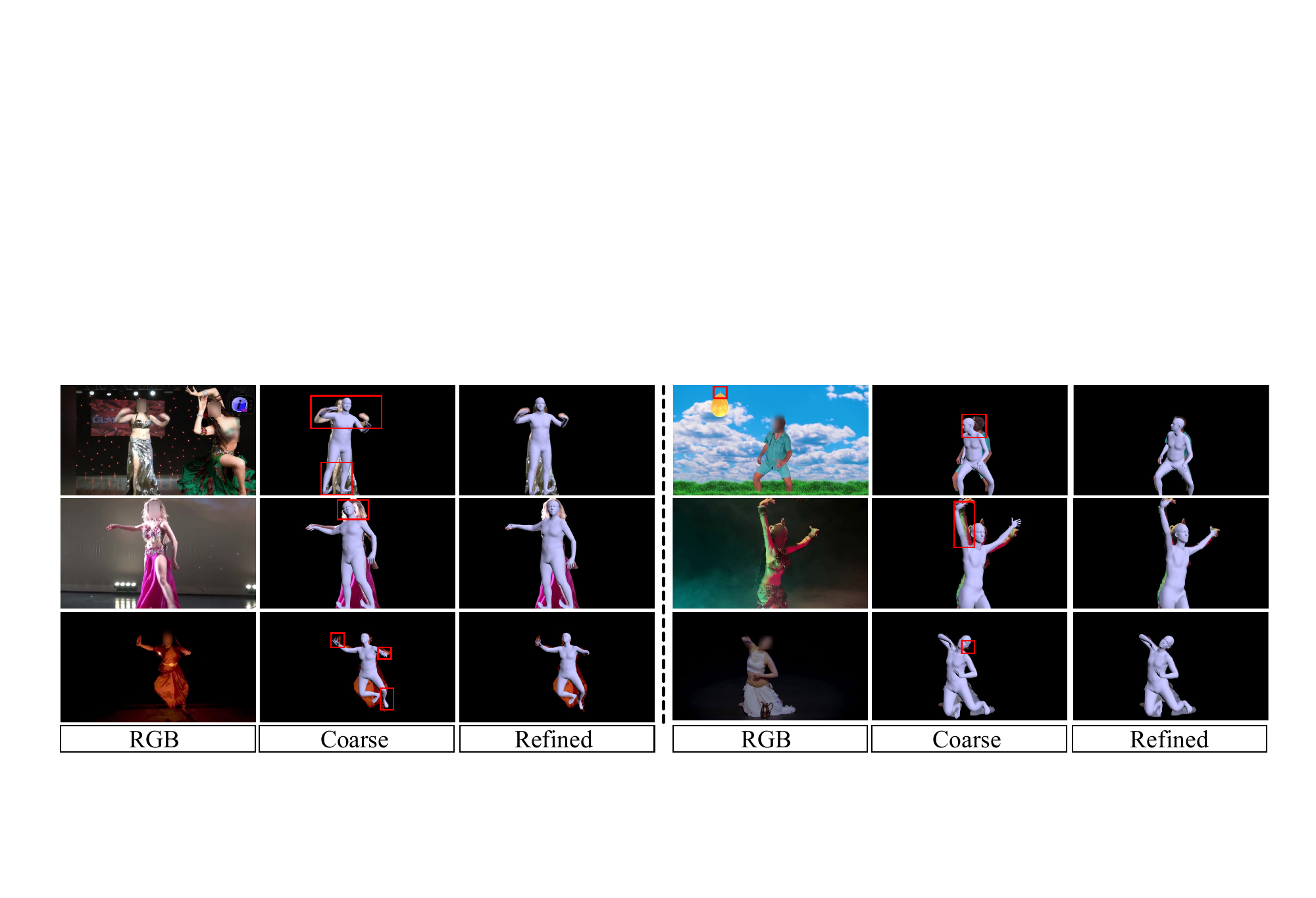}
    \caption{Comparison of the coarse and refined SMPL parameters. The coarse SMPL annotations are from Stage III, and are later refined in Stage IV. The refined SMPL parameters achieve better alignments to the raw RGB images.}
    \label{fig:appendix-coarse-refined-comparison}
    \vspace{-10pt}
\end{figure*}

We adopt the loss functions as mentioned above for supervision:
\begin{equation}
    \begin{aligned}
    \mathcal{L}=
    \lambda^{s}_\theta \mathcal{L}^{s}_\theta +
    \lambda^{s}_R \mathcal{L}^{s}_R +
    \lambda^{s}_T \mathcal{L}^{s}_T + 
    \lambda^{s}_{2D} \mathcal{L}^{s}_{2D} \\+
    \lambda_{2D} \mathcal{L}_{2D} +
    \lambda_{\mathrm{mask}} \mathcal{L}_{\mathrm{mask}} +
    \lambda_{\mathrm{prior}} \mathcal{L}_{\mathrm{prior}}.
    \end{aligned}
\end{equation}
We empirically set the loss weights as $\lambda^{s}_\theta=100$, $\lambda^{s}_R=1000$, $\lambda^{s}_T=50$, $\lambda^{s}_{2D}=100$, $\lambda_{2D}=100$, $\lambda_{\mathrm{mask}}=100$ and $\lambda_{\mathrm{prior}}=0.1$. We adopt the LBFGS~\cite{LBFGS} optimizer with the learning rate $lr=1.0$. 
Qualitative comparisons between coarse and refined SMPL and camera parameters can be found in~\Cref{fig:appendix-coarse-refined-comparison}. Stage IV boosts local alignments on extremities.

\begin{figure*}[!h]
  \centering    \includegraphics[width=0.99\linewidth]{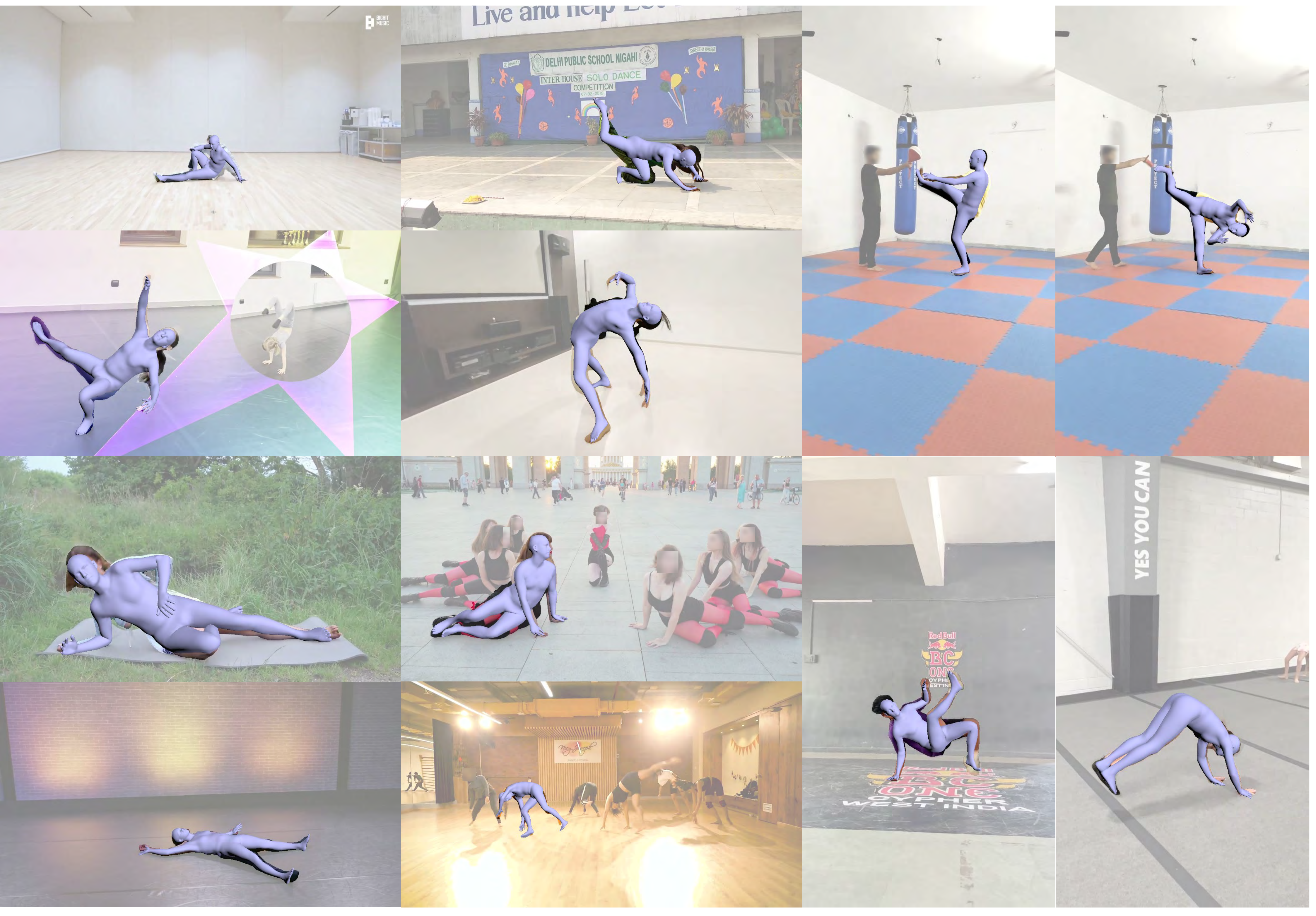}
    \caption{Visualization of SMPL overlay on unusual camera viewpoints. Our pipeline can show robust annotations on unusual camera viewpoints and body poses. }
    \label{fig:unusual}
    \vspace{-5pt}
\end{figure*}

\noindent\textbf{Efficiency.} 
Data collection and annotation efficiency are also crucial for data scale-up and application. Despite the complex multiple-stage design, our data-collecting pipeline only takes less than $200$ seconds to generate full annotations for a $20$ seconds in-the-wild video clip.

\section{Details in Pipeline Evaluation}
\label{sec:appendix-details-in-pipeline-evaluation}
\noindent\textbf{Dataset.}
We evaluate our four-stage pipeline on the in-the-wild EMDB~\cite{EMDB}, which is widely recognized for its challenging and diverse real-world scenarios. We use the EMDB-1 split to evaluate the camera-coordinate performance. EMDB-1 contains $17$ sequences totaling $13.5$ minutes. 

\noindent\textbf{Metrics.}
For joints, we compute error on the $24$ main joints of the human body under the SMPL convention. As for vertex, we calculate the point-to-point corresponding error on the SMPL vertices. 
We report quantitative results on MPJPE, PA-MPJPE~\cite{h36m}, and PVE~\cite{XueCZYMM22}. 
\noindent\textbf{\emph{MPJPE (Mean Per Joint Position Error)}} calculates the mean distances between the predicted and ground-truth 3D joints after the translation alignment at the pelvis joint. The predicted or ground-truth 3D joints are regressed from corresponding pose and shape parameters.
\noindent\textbf{\emph{PA-MPJPE (Procrustes analysis MPJPE)}} calculates the mean distances between the predicted and ground-truth 3D joints after Procrustes Analysis~\cite{PA}, including alignment in scale, translation and rotation. PA-MPJPE mainly focuses on the quality of pose and shape estimation, regardless of global rotation.
\noindent\textbf{\emph{PVE (Per Vertex Error)}} calculates the mean distances between the vertices on the human mesh without any alignment, which evaluates the reconstruction accuracy of the human surface. 

\noindent\textbf{Qualitative Comparisons.}
Qualitative comparisons on the EMDB dataset are also shown in~\Cref{fig:qualitative-comparison} (a)~\textasciitilde~(d). Compared to previous state-of-the-art HMR2.0~\cite{human-in-4d}, our pipeline can adapt to complex environmental scenes and output reasonable results in uncommon scenarios. 
For example, our pipeline can accurately predict 1) the foot and ankle pose of~\Cref{fig:qualitative-comparison} ({a}) \& ({c}); 2) the head and neck pose of~\Cref{fig:qualitative-comparison} ({b}) \& ({d}); 3) the global alignment of~\Cref{fig:qualitative-comparison} ({a}), ({b}) \& ({d}). 
Qualitative comparisons on the 3DPW dataset are also shown in~\Cref{fig:qualitative-comparison} (e)~\textasciitilde~(h). Compared to the official ground truth (GT) from expensive IMUs, our annotations also exhibit better alignment on 1) the foot and ankle pose of~\Cref{fig:qualitative-comparison} (e)~\textasciitilde~(g); 2) the head and neck pose of~\Cref{fig:qualitative-comparison} (g); 3) the hand pose of~\Cref{fig:qualitative-comparison} (h).
These comparisons validate the annotation quality of our pipeline for in-the-wild videos.

\section{License, Statistics and Visualizations}
\begin{figure*}[!h]
  \centering    \includegraphics[width=0.99\linewidth]{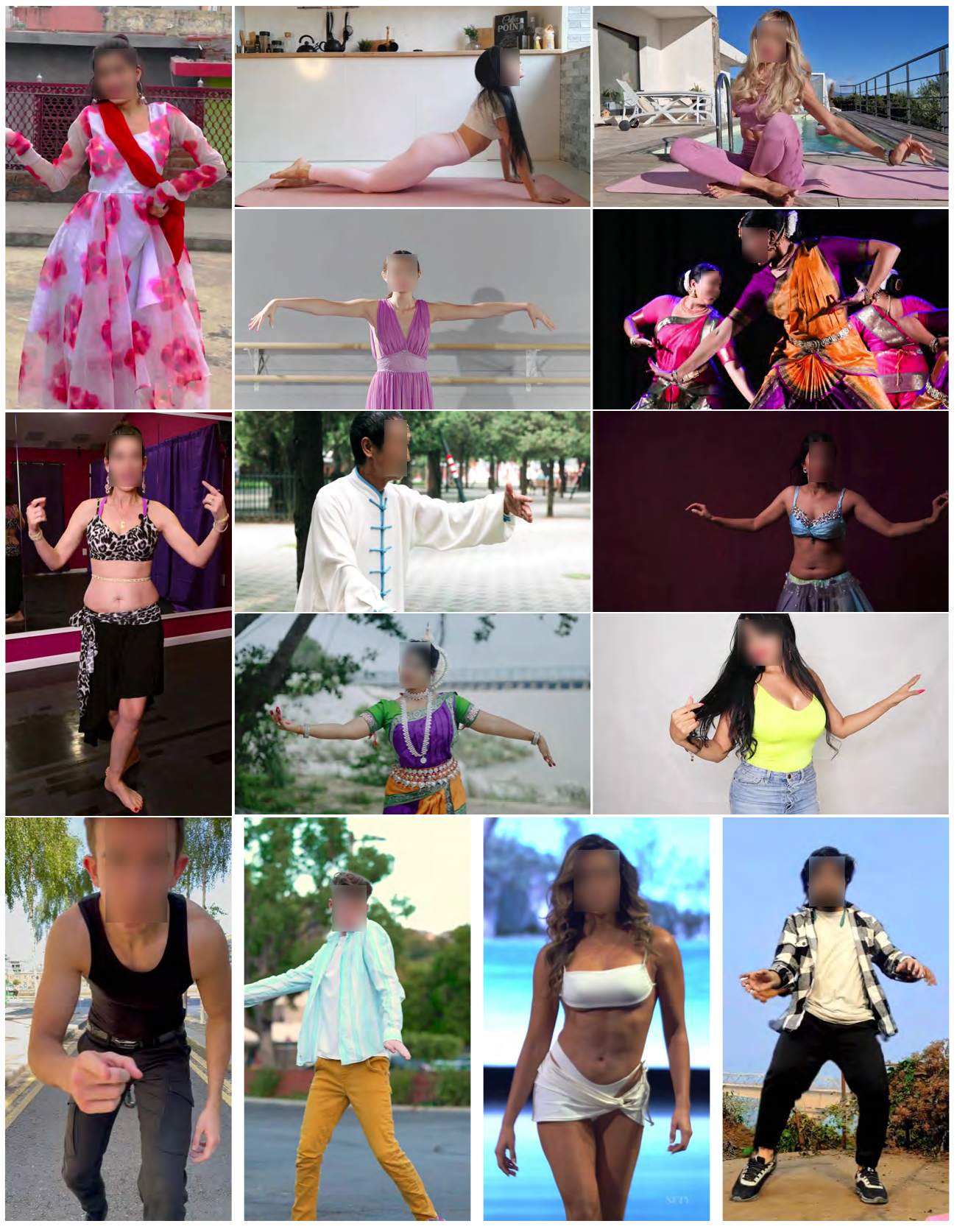}
    \caption{More RGB examples from the proposed WildAvatar dataset. The best view zoomed in on-screen for details.}
    \label{fig:hq}
\end{figure*}

\begin{figure*}[!h]
  \centering    \includegraphics[width=0.7\linewidth]{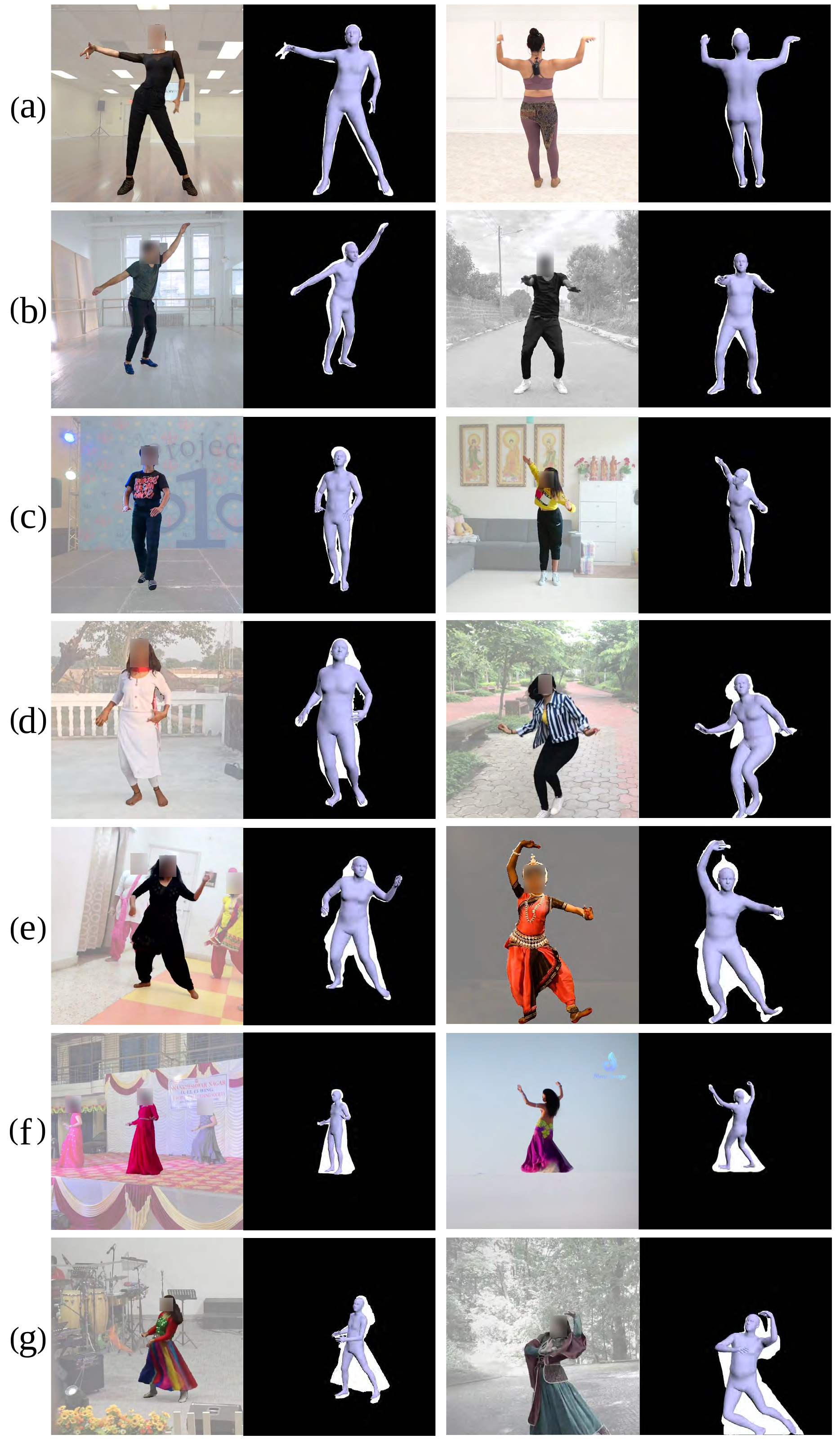}
    \caption{Examples of different SSIOU ranges. SSIOU rises from up to down. The last row shows the SSIOU larger than $1.9$. The samples with SSIOU larger than $1.9$ are mostly caused by loose dresses rather than erroneous SMPL fitting.}
    \label{fig:ssiou}
\end{figure*}

The authors bear all responsibility in case of violation of rights and confirm that this dataset is open-sourced under the \textbf{S-Lab License 1.0 license}. We shall enforce strict regulations when applying our code and data to mitigate potential negative social impacts. $69.1\%$ / $30.9\%$ scenes in WildAvatar are indoor/outdoor, respectively. $45.3\%$ / $54.7\%$ of the scenes in WildAvatar have single/multiple human(s), respectively. And $34.6\%$ / $65.4\%$ subjects in WildAvatar are male/female, respectively. More visualization of SMPL overlay on unusual camera viewpoints can be found in~\Cref{fig:unusual}. More RGB examples from the proposed WildAvatar dataset can be found in~\Cref{fig:hq}, and examples of different SSIOU ranges can be found in~\Cref{fig:ssiou}.

